\pdfoutput=1

\documentclass[11pt]{article}
\usepackage{CJKutf8}

\usepackage[]{emnlp2021}

\usepackage{times}
\usepackage{latexsym}

\usepackage[T1]{fontenc}

\usepackage[utf8]{inputenc}

\usepackage{microtype}

\usepackage{tabularx}
\usepackage{booktabs}
\usepackage{tikz}
\usepackage{pgfplots}
\usepackage{amsmath}
\usepackage{amssymb}
\usepackage{graphicx}
\usepackage{multirow}
\usepackage{subcaption}
\usepackage{pifont}
\usepackage{spverbatim}

\newif\ifshowcomments
\showcommentstrue 

\newcommand{\ours}[0]{Toolformer}

\newcommand{\cmark}{\ding{51}}%
\newcommand{\xmark}{\ding{55}}%

\pgfplotsset{compat=1.13}

\definecolor{c0}{cmyk}{1,0.3968,0,0.2588} 
\definecolor{c1}{cmyk}{0,0.6175,0.8848,0.1490} 
\definecolor{c2}{cmyk}{0.1127,0.6690,0,0.4431} 
\definecolor{c3}{cmyk}{0.3081,0,0.7209,0.3255} 
\definecolor{c4}{cmyk}{0.6765,0.2017,0,0.0667} 
\definecolor{c5}{cmyk}{0,0.8765,0.7099,0.3647} 

\definecolor{c0alt}{RGB}{15,158,251} 

\definecolor{darkgrey}{RGB}{149,149,149}
\definecolor{decentgrey}{RGB}{242,242,242}

\usetikzlibrary{calc,fit,positioning,arrows,arrows.meta,backgrounds,decorations.pathreplacing}
\usepgfplotslibrary{fillbetween}

\title{Toolformer: Language Models Can Teach Themselves to Use Tools}

\author{Timo Schick \quad Jane Dwivedi-Yu\quad Roberto Dess\`i$^\dagger$ \quad Roberta Raileanu \\[4pt]
\bf Maria Lomeli \quad Luke Zettlemoyer \quad Nicola Cancedda \quad Thomas Scialom
\\[8pt]
Meta AI Research \
$^{\dagger}$Universitat Pompeu Fabra
}

\begin{document}
\begin{CJK*}{UTF8}{gbsn}
\maketitle

\begin{abstract} 
Language models (LMs) exhibit remarkable abilities to solve new tasks from just a few examples or textual instructions, especially at scale. They also, paradoxically, struggle with basic functionality, such as arithmetic or factual lookup, where much simpler and smaller models excel. In this paper, we show that LMs can teach themselves to \emph{use external tools} via simple APIs and achieve the best of both worlds. We introduce \emph{\ours{}}, a model trained to decide which APIs to call, when to call them, what arguments to pass, and how to best incorporate the results into future token prediction. This is done in a self-supervised way, requiring nothing more than a handful of demonstrations for each API.
We incorporate a range of tools, including a calculator, a Q\&A system, a search engine, a translation system, and a calendar. 
\ours{} achieves substantially improved zero-shot performance across a variety of downstream tasks, often competitive with much larger models, without sacrificing its core language modeling abilities.
\end{abstract}

\section{Introduction}

Large language models achieve impressive zero- and few-shot results on a variety of natural language processing tasks \citep[][i.a.]{brown2020language,chowdhery2022palm} and show several emergent capabilities \citep{wei2022emergent}. However, all of these models have several inherent limitations that can at best be partially addressed by further scaling. These limitations include an inability to access up-to-date information on recent events \citep{komeili-etal-2022-internet} and the related tendency to hallucinate facts \citep{maynez2020faithfulness,ji2022survey}, difficulties in understanding low-resource languages \citep{lin2021fewshot}, a lack of mathematical skills to perform precise calculations \citep{patel-etal-2021-nlp} and an unawareness of the progression of time \citep{dhingra-etal-2022-time}.

\begin{figure}
    \centering
    \includegraphics[width=\linewidth]{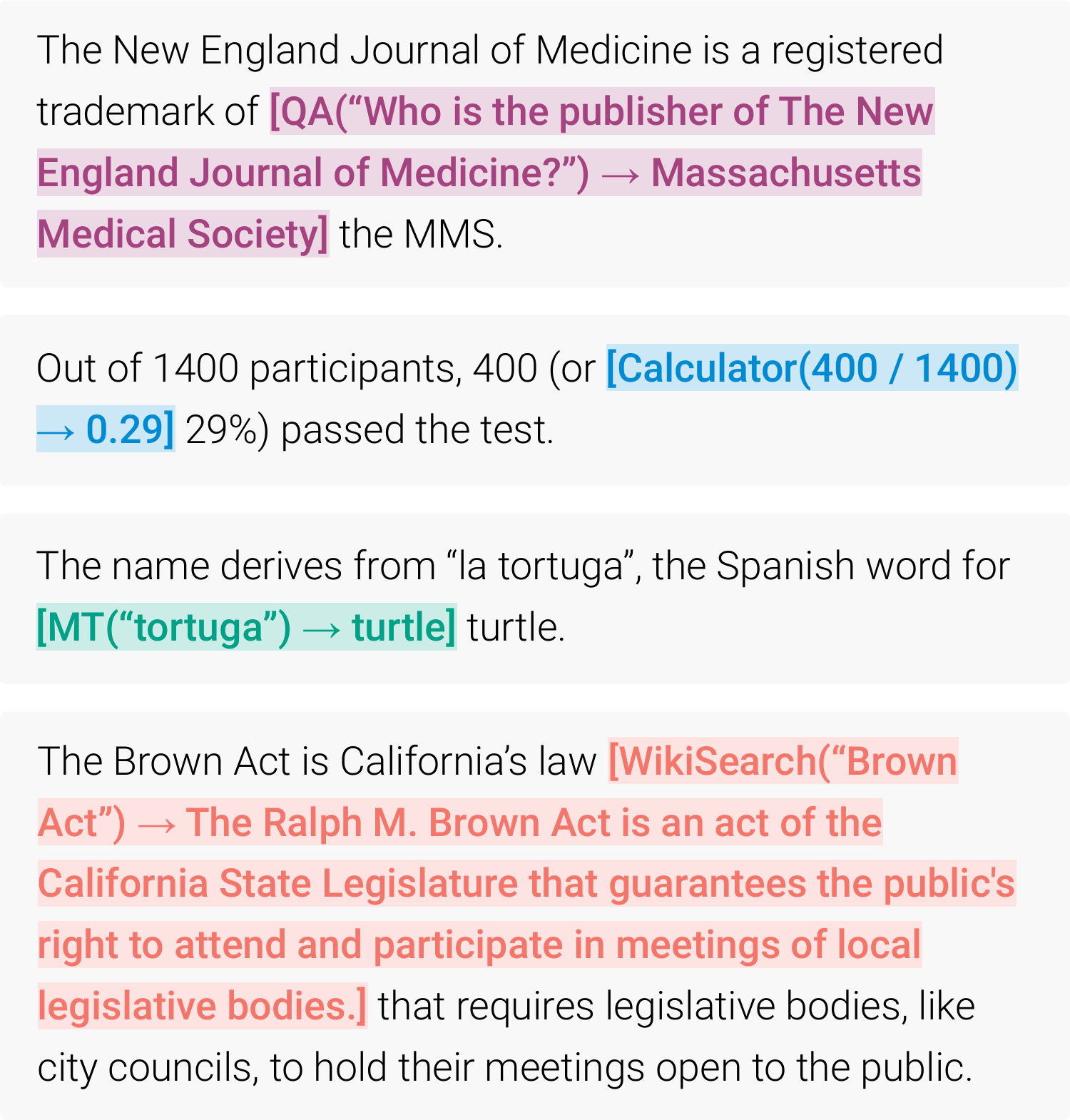}
    \caption{Exemplary predictions of \ours{}. The model autonomously decides to call different APIs (from top to bottom: a question answering system, a calculator, a machine translation system, and a Wikipedia search engine) to obtain information that is useful for completing a piece of text.}
    \label{fig:example}
\end{figure}

\begin{figure*}
    \centering
    \includegraphics[width=\linewidth]{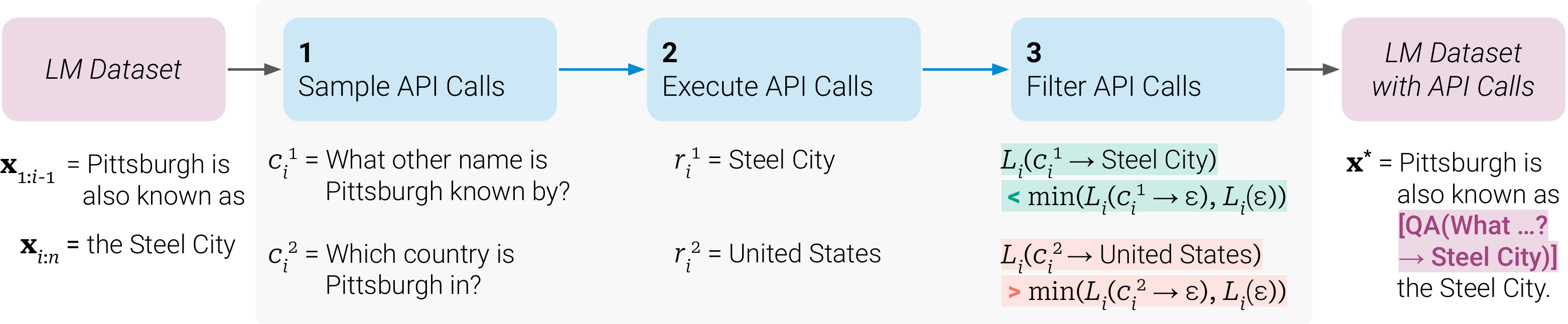}
    \caption{Key steps in our approach, illustrated for a \emph{question answering} tool: Given an input text $\mathbf{x}$, we first sample a position $i$ and corresponding API call candidates $c_i^1, c_i^2, \ldots, c_i^k$. We then execute these API calls and filter out all calls which do not reduce the loss $L_i$ over the next tokens. All remaining API calls are interleaved with the original text, resulting in a new text $\mathbf{x}^*$.}
    \label{fig:approach}
\end{figure*}

A simple way to overcome these limitations of today's language models is to give them the ability to \emph{use external tools} such as search engines, calculators, or calendars.
However, existing approaches either rely on large amounts of human annotations \citep{komeili-etal-2022-internet,thoppilan2022lamda} or limit tool use to task-specific settings only \citep[e.g.,][]{gao2022pal,parisi2022talm}, hindering a more widespread adoption of tool use in LMs.
Therefore, we propose \emph{\ours{}}, a model that learns to use tools in a novel way, which fulfills the following desiderata:
\begin{itemize}
    \item The use of tools should be learned in a self-supervised way without requiring large amounts of \emph{human annotations}. This is important not only because of the costs associated with such annotations, but also because what humans find useful may be different from what a model finds useful.
    \item The LM should not lose any of its \emph{generality} and should be able to decide for itself \emph{when} and \emph{how} to use which tool. In contrast to existing approaches, this enables a much more comprehensive use of tools that is not tied to specific tasks.
\end{itemize}

Our approach for achieving these goals is based on the recent idea of using large LMs with \emph{in-context learning} \citep{brown2020language} to generate entire datasets from scratch \citep{schick-schutze-2021-generating,honovich2022unnatural,wang2022selfinstruct}: Given just a handful of human-written examples of how an API can be used, we let a LM annotate a huge language modeling dataset with potential API calls. We then use a self-supervised loss to determine which of these API calls actually help the model in predicting future tokens. Finally, we finetune the LM itself on the API calls that it considers useful. As illustrated in Figure~\ref{fig:example}, through this simple approach, LMs can learn to control a variety of tools, and to choose for themselves which tool to use when and how. 

As our approach is agnostic of the dataset being used, we can apply it to the exact same dataset that was used to pretrain a model in the first place. This ensures that the model does not lose any of its generality and language modeling abilities. We conduct experiments on a variety of different downstream tasks, demonstrating that after learning to use tools, \ours{}, which is based on a pretrained GPT-J model \citep{gpt-j} with 6.7B parameters, achieves much stronger zero-shot results, clearly outperforming a much larger GPT-3 model \citep{brown2020language} and several other baselines on various tasks.

\section{Approach}
\label{section:approach}

Our aim is to equip a language model $M$ with the ability to use different tools by means of API calls. We require that inputs and outputs for each API can be represented as text sequences.
This allows seamless insertion of API calls into any given text, using special tokens to mark the start and end of each such call. 

We represent each API call as a tuple $c = ({a}_c, {i}_c)$ where $a_c$ is the name of the API and $i_c$ is the corresponding input. Given an API call $c$ with a corresponding result $r$, we denote the linearized sequences of the API call not including and including its result, respectively, as:
\begin{align*}
\text{e}(c)     & = \texttt{<API>}\, a_c \texttt{(} i_c \texttt{)}\, \texttt{</API>} \\
\text{e}(c, r)  & = \texttt{<API>}\, a_c \texttt{(} i_c \texttt{)} \rightarrow r\, \texttt{</API>}
\end{align*}
 where ``\texttt{<API>}'', ``\texttt{</API>}'' and ``$\rightarrow$'' are special tokens.\footnote{In practice, we use the token sequences ``\texttt{ [}'', ``\texttt{]}'' and ``\texttt{->}'' to represent ``\texttt{<API>}'', ``\texttt{</API>}'' and ``$\rightarrow$'', respectively. This enables our approach to work without modifying the existing LM's vocabulary. For reasons of readability, we still refer to them as ``\texttt{<API>}'', ``\texttt{</API>}'' and ``$\rightarrow$'' throughout this section.} Some examples of linearized API calls inserted into text sequences are shown in Figure~\ref{fig:example}.

Given a dataset $\mathcal{C} = \{ \mathbf{x}^1, \ldots, \mathbf{x}^{|\mathcal{C}|} \}$ of plain texts, we first convert this dataset into a dataset $\mathcal{C}^*$ augmented with API calls. This is done in three steps, illustrated in Figure~\ref{fig:approach}: First, we exploit the in-context learning ability of $M$ to sample a large number of potential API calls. We then execute these API calls and finally check whether the obtained responses are helpful for predicting future tokens; this is used as a filtering criterion. After filtering, we merge API calls for different tools, resulting in the augmented dataset $\mathcal{C}^*$, and finetune $M$ itself on this dataset. Each of these steps is described in more detail below.

\paragraph{Sampling API Calls} For each API, we write a prompt $P(\mathbf{x})$ that encourages the LM to annotate an example $\mathbf{x} = x_1, \ldots, x_n$ with API calls. An example of such a prompt for a question answering tool is shown in Figure~\ref{fig:api_call_prompt}; all prompts used are shown in Appendix~\ref{appendix:tool-prompts}. Let $p_M(z_{n+1} \mid z_1, \ldots, z_n)$ be the probability that $M$ assigns to token $z_{n+1}$ as a continuation for the sequence $z_1, \ldots, z_n$. We first sample up to $k$ candidate \emph{positions} for doing API calls by computing, for each $i \in \{1, \ldots, n\}$, the probability
\[
p_i = p_M(\texttt{<API>} \mid P(\mathbf{x}), x_{1:i-1} )
\]
that $M$ assigns to starting an API call at position $i$. 
Given a sampling threshold $\tau_s$, we keep all positions $I = \{ i \mid p_i > \tau_s \}$; if there are more than $k$ such positions, we only keep the top $k$.

For each position $i \in I$, we then obtain up to $m$ API calls $c_i^1, \ldots, c_i^m$ by sampling from $M$ given the sequence $[P(\mathbf{x}), x_1, \ldots, x_{i-1}, \texttt{<API>}]$ as a prefix and $\texttt{</API>}$ as an end-of-sequence token.\footnote{We discard all examples where $M$ does not generate the $\texttt{</API>}$ token.} 

\begin{figure}
    \centering
    \includegraphics[width=\linewidth]{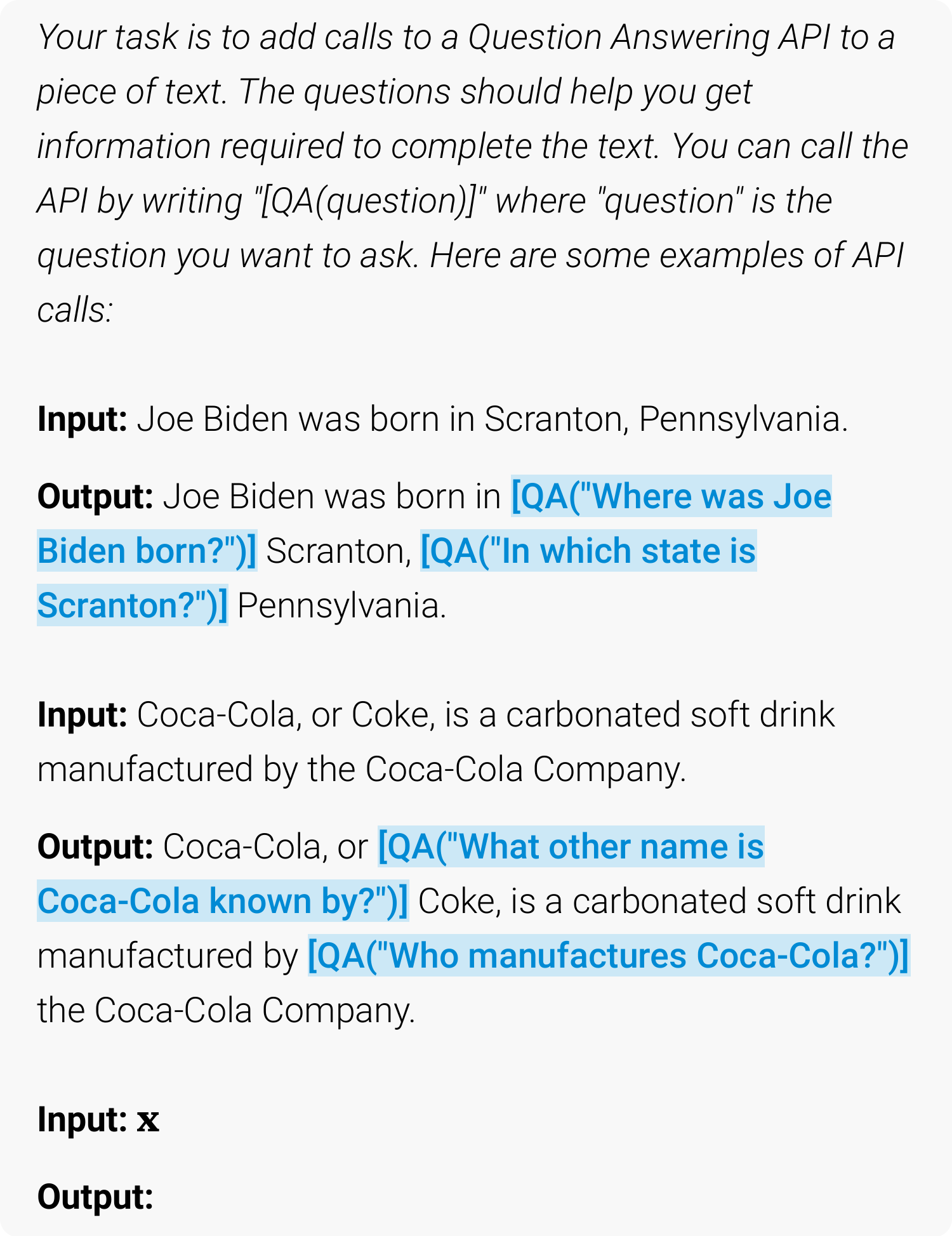}
    \caption{An exemplary prompt $P(\mathbf{x})$ used to generate API calls for the question answering tool.}
    \label{fig:api_call_prompt}
\end{figure}

\paragraph{Executing API Calls} As a next step, we execute all API calls generated by $M$ to obtain the corresponding results. How this is done depends entirely on the API itself -- for example, it can involve calling another neural network, executing a Python script or using a retrieval system to perform search over a large corpus. The response for each API call $c_i$ needs to be a single text sequence $r_i$.

\paragraph{Filtering API Calls} Let $i$ be the position of the API call $c_i$ in the sequence $\mathbf{x} = x_1, \ldots, x_n$, and let $r_i$ be the response from the API. Further, given a sequence $(w_i \mid i \in \mathbb{N})$ of \emph{weights}, let 
\[
L_i(\mathbf{z}) = -\sum_{j=i}^n w_{j-i} \cdot \log {p_M(x_j \mid \mathbf{z}, x_{1:j-1})} 
\]
be the weighted cross entropy loss for $M$ over the tokens $x_i, \ldots, x_n$ if the model is prefixed with $\mathbf{z}$.
We compare two different instantiations of this loss:
\begin{align*}
L_i^+ & = L_i(\text{e}(c_i, r_i))\\
L_i^- & =  \min \left( L_i(\varepsilon), L_i(\text{e}(c_i, \varepsilon )) \right)
\end{align*}
where $\varepsilon$ denotes an empty sequence. The former is the weighted loss over all tokens $x_i, \ldots, x_n$ if the API call and its result are given to $M$ as a prefix;\footnote{We provide $\text{e}(c_i, r_i)$ as a prefix instead of inserting it at position $i$ because $M$ is not yet finetuned on any examples containing API calls, so inserting it in the middle of $\mathbf{x}$ would interrupt the flow and not align with patterns in the pretraining corpus, thus hurting perplexity.} the latter is the minimum of the losses obtained from (i) doing no API call at all and (ii) doing an API call, but not providing the response. Intuitively, an API call is helpful to $M$ if providing it with both the input \emph{and} the output of this call makes it easier for the model to predict future tokens, compared to not receiving the API call at all, or receiving only its input. Given a filtering threshold $\tau_f$, we thus only keep API calls for which
\[
L_i^- - L_i^+  \geq \tau_f
\]
holds, i.e., adding the API call and its result \emph{reduces} the loss by at least $\tau_f$, compared to not doing any API call or obtaining no result from it. 

\paragraph{Model Finetuning} After sampling and filtering calls for all APIs, we finally merge the remaining API calls and interleave them with the original inputs. That is, for an input text $\mathbf{x} = x_1, \ldots, x_n$ with a corresponding API call and result $(c_i, r_i)$ at position $i$, we construct the new sequence $\mathbf{x}^* = x_{1:{i-1}}, \text{e}(c_i, r_i), x_{i:n}$; we proceed analogously for texts with multiple API calls. Doing this for all $\mathbf{x} \in \mathcal{C}$ results in the new dataset $\mathcal{C}^*$ augmented with API calls. We use this new dataset to finetune $M$, using a standard language modeling objective. Crucially,  apart from inserted API calls the augmented dataset $\mathcal{C}^*$ contains the exact same texts as $\mathcal{C}$, the original dataset. As a consequence, finetuning $M$ on $\mathcal{C}^*$ exposes it to the same content as finetuning on $\mathcal{C}$. Moreover, as API calls are inserted in exactly those positions and with exactly those inputs that help $M$ predict future tokens, finetuning on $\mathcal{C}^*$ enables the language model to decide when and how to use which tool, based purely on its own feedback.

\paragraph{Inference} When generating text with $M$ after finetuning with our approach, we perform regular decoding until $M$ produces the ``$\rightarrow$'' token, indicating that it next expects the response for an API call. At this point, we interrupt the decoding process, call the appropriate API to get a response, and continue the decoding process after inserting both the response and the $\texttt{</API>}$ token.

\section{Tools}

We explore a variety of tools to address different shortcomings of regular LMs. The only constraints we impose on these tools is that (i) both their inputs and outputs can be represented as text sequences, and (ii) we can obtain a few demonstrations of their intended use. Concretely, we explore the following five tools: a question answering system, a Wikipedia search engine, a calculator, a calendar, and a machine translation system. Some examples of potential calls and return strings for the APIs associated with each of these tools are shown in Table~\ref{tab:tool-examples}. We briefly discuss all tools below; further details can be found in Appendix~\ref{appendix:api-details}.

\begin{table*}
\small
\begin{tabularx}{\linewidth}{lp{3cm}X}
\toprule
\textbf{API Name} & \textbf{Example Input} & \textbf{Example Output} \\
\midrule
Question Answering & Where was the Knights of Columbus founded? & New Haven, Connecticut \\\addlinespace[0.2cm]
Wikipedia Search & Fishing Reel Types & Spin fishing > Spin fishing is distinguished between fly fishing and bait cast fishing by the type of rod and reel used. There are two types of reels used when spin fishing, the open faced reel and the closed faced reel. \\\addlinespace[0.2cm]
Calculator & 27 + 4 * 2 & 35 \\\addlinespace[0.2cm]
Calendar & $\varepsilon$ & Today is Monday, January 30, 2023. \\\addlinespace[0.2cm]
Machine Translation & sûreté nucléaire & nuclear safety \\
\bottomrule
\end{tabularx}
\caption{Examples of inputs and outputs for all APIs used.}
\label{tab:tool-examples}
\end{table*}

\paragraph{Question Answering} Our first tool is a question answering system based on another LM that can answer simple factoid questions. Specifically, we use \emph{Atlas} \citep{izacard2022atlas}, a retrieval-augmented LM finetuned on Natural Questions \citep{kwiatkowski-etal-2019-natural}.

\paragraph{Calculator} As a second tool, we use a calculator that can perform simple numeric calculations; we only support the four basic arithmetic operations. Results are always rounded to two decimal places.

\paragraph{Wikipedia Search} Our third tool is a search engine that, given a search term, returns short text snippets from Wikipedia. Compared to our question answering tool, this search enables a model to get more comprehensive information on a subject, but requires it to extract the relevant parts by itself. As our search engine, we use a BM25 retriever \citep{robertson1995okapi,baeza1999modern} that indexes the Wikipedia dump from KILT \citep{petroni-etal-2021-kilt}.

\paragraph{Machine Translation System} Our fourth tool is a machine translation system based on a LM that can translate a phrase from any language into English. More concretely, we use the 600M parameter NLLB~\citep{costa2022no} as our multilingual machine translation model that works for 200 languages (including low-resource ones). The source language is automatically detected using the \textit{fastText} classifier~\citep{joulin2016fasttext}, while the target language is always set to English. 

\paragraph{Calendar}
Our final tool is a calendar API that, when queried, returns the current date without taking any input. This provides temporal context for predictions that require some awareness of time.

\section{Experiments}

We investigate whether our approach enables a model to use tools without any further supervision and to decide for itself when and how to call which of the available tools. To test this, we select a variety of downstream tasks where we assume at least one of the considered tools to be useful, and evaluate performance in zero-shot settings (Section~\ref{section:downstream-tasks}). Beyond that, we also ensure that our approach does not hurt the model's core language modeling abilities; we verify this by looking at perplexity on two language modeling datasets (Section~\ref{section:language-modeling}). Finally, we investigate how the ability to learn using tools is affected by model size (Section~\ref{section:scaling-laws}).

\subsection{Experimental Setup} 
\label{section:experimental-setup}

\paragraph{Dataset Generation}
Throughout all of our experiments, we use a subset of CCNet \citep{wenzek-etal-2020-ccnet} as our language modeling dataset $\mathcal{C}$ and GPT-J \citep{gpt-j} as our language model $M$. To reduce the computational cost of annotating $\mathcal{C}$ with API calls, we define heuristics for some APIs to get a subset of $\mathcal{C}$ for which API calls are more likely to be helpful than for an average text. For example, we only consider texts for the calculator tool if they contain at least three numbers.  Details of the heuristics used are given in Appendix~\ref{appendix:api-details}. For obtaining $\mathcal{C}^*$ from $\mathcal{C}$, we perform all steps described in Section~\ref{section:approach} and additionally filter out all examples for which all API calls were eliminated in the filtering step.\footnote{While this filtering alters the distribution of training examples, we assume that the remaining examples are close enough to the original distribution so that $M$'s language modeling abilities remain unaffected. This assumption is empirically validated in Section~\ref{section:language-modeling}.} For the weighting function, we use 
\[
w_t  = \frac{\tilde{w}_t}{ \sum_{s \in \mathbb{N}} \tilde{w}_s} \text{ with }
\tilde{w}_t = \max(0, 1 - 0.2 \cdot t)
\]
to make sure that API calls happen close to where the information provided by the API is actually helpful for the model. The thresholds $\tau_s$ and $\tau_f$ are chosen individually for each tool to ensure a sufficiently larger number of examples; see Appendix~\ref{appendix:api-details} for details.
Table~\ref{tab:c_star} shows relevant statistics of our final dataset augmented with API calls. 

\begin{table}
    \centering
    \small
    \setlength{\tabcolsep}{5pt}
    \begin{tabularx}{\linewidth}{Xccc}
         \toprule
         & \multicolumn{3}{c}{\textbf{Number of Examples}} \\
         \textbf{API} & $\tau_f = 0.5$ & $\tau_f = 1.0$ & $\tau_f = 2.0$ \\
         \midrule
         Question Answering & \phantom{0}51,987 & \phantom{0}18,526 & \phantom{00}5,135 \\
         Wikipedia Search & 207,241 & \phantom{0}60,974 & \phantom{0}13,944 \\
         Calculator & \phantom{00}3,680 & \phantom{000,}994 & \phantom{000,}138 \\
         Calendar & \phantom{0}61,811 & \phantom{0}20,587 & \phantom{00}3,007 \\
         Machine Translation & \phantom{00}3,156 & \phantom{00}1,034 &  \phantom{000,}229 \\
         \bottomrule
    \end{tabularx}
    \caption{Number of examples with API calls in $\mathcal{C}^*$ for different values of our filtering threshold $\tau_f$.}
    \label{tab:c_star}
\end{table}

\paragraph{Model Finetuning}
We finetune $M$ on $\mathcal{C}^*$ using a batch size of 128 and a learning rate of $1\cdot10^{-5}$ with linear warmup for the first 10\% of training. Details of our finetuning procedure are given in Appendix~\ref{appendix:finetuning}. 

\paragraph{Baseline Models}
Throughout the remainder of this section, we mainly compare the following models:

\begin{itemize}
\item \textbf{GPT-J}: A regular GPT-J model without any finetuning.
\item \textbf{GPT-J + CC}: GPT-J finetuned on $\mathcal{C}$, our subset of CCNet \emph{without} any API calls.
\item \textbf{\ours{}}: GPT-J finetuned on $\mathcal{C}^*$, our subset of CCNet augmented with API calls.
\item \textbf{\ours{} (disabled)}: The same model as \ours{}, but API calls are disabled during decoding.\footnote{This is achieved by manually setting the probability of the $\texttt{<API>}$ token to 0.}
\end{itemize}
For most tasks, we additionally compare to OPT (66B) \citep{zhang2022opt} and GPT-3\footnote{We use the original \texttt{davinci} variant that is not finetuned on any instructions.} (175B) \citep{brown2020language}, two models that are about 10 and 25 times larger than our other baseline models, respectively.

\subsection{Downstream Tasks}
\label{section:downstream-tasks}

We evaluate all models on a variety of downstream tasks. In all cases, we consider a prompted zero-shot setup -- i.e., models are instructed to solve each task in natural language, but we do not provide any in-context examples. This is in contrast to prior work on tool use \citep[e.g.,][]{gao2022pal,parisi2022talm}, where models are provided with dataset-specific examples of how a tool can be used to solve a concrete task. We choose the more challenging zero-shot setup as we are interested in seeing whether \ours{} works in precisely those cases where a user does not specify in advance which tools should be used in which way for solving a specific problem.

We use standard greedy decoding, but with one modification for \ours{}: We let the model start an API call not just when \texttt{<API>} is the most likely token, but whenever it is one of the $k$ most likely tokens. For $k = 1$, this corresponds to regular greedy decoding; we instead use $k = 10$ to increase the disposition of our model to make use of the APIs that it has access to. At the same time, we only at most one API call per input to make sure the model does not get stuck in a loop where it constantly calls APIs without producing any actual output. The effect of these modifications is explored in Section~\ref{section:analysis}.

\subsubsection{LAMA}

We evaluate our models on the SQuAD, Google-RE and T-REx subsets of the LAMA benchmark \citep{petroni-etal-2019-language}. For each of these subsets, the task is to complete a short statement with a missing fact (e.g., a date or a place). 
As LAMA was originally designed to evaluate \emph{masked} language models \citep[e.g.,][]{devlin-etal-2019-bert}, we filter out examples where the mask token is not the final token, so that the remaining examples can be processed in a left-to-right fashion. To account for different tokenizations and added complexity from not informing the model that a single word is required, we use a slightly more lenient evaluation criterion than exact match and simply check whether the correct word is within the first five words predicted by the model. As LAMA is based on statements obtained directly from Wikipedia, we prevent \ours{} from using the Wikipedia Search API to avoid giving it an unfair advantage. 

Results for all models can be seen in Table~\ref{tab:lama_results}. All GPT-J models without tool use achieve similar performance. Crucially, \ours{} clearly outperforms these baseline models, improving upon the best baseline by 11.7, 5.2 and 18.6 points, respectively. It also clearly outperforms OPT (66B) and GPT-3 (175B), despite both models being much larger. This is achieved because the model independently decides to ask the question answering tool for the required information in almost all cases (98.1\%); for only very few examples, it uses a different tool (0.7\%) or no tool at all (1.2\%).

\begin{table}
    \centering
    \small
    \setlength{\tabcolsep}{3pt}
    \begin{tabularx}{\linewidth}{Xccc}
        \toprule
        \textbf{Model} & \textbf{SQuAD} & \textbf{Google-RE} & \textbf{T-REx} \\
        \midrule
         GPT-J & 17.8 & \phantom{0}4.9 & 31.9 \\         
         GPT-J + CC & 19.2 & \phantom{0}5.6 & 33.2 \\        
         \ours{} (disabled) & 22.1 & \phantom{0}6.3 & 34.9  \\
        \ours{} & \underline{\textbf{33.8}} & \underline{\textbf{11.5}} & \underline{\textbf{53.5}} \\
         \midrule
         OPT (66B) & 21.6 & \phantom{0}2.9 & 30.1 \\
         GPT-3 (175B) & 26.8 & \phantom{0}7.0 & 39.8 \\
         \bottomrule
    \end{tabularx}
    \caption{Results on subsets of LAMA. \ours{} uses the question answering tool for most examples, clearly outperforming all baselines of the same size and achieving results competitive with GPT-3 (175B).}
    \label{tab:lama_results}
\end{table}

\subsubsection{Math Datasets}

We test mathematical reasoning abilities on ASDiv \citep{miao-etal-2020-diverse}, SVAMP \citep{patel-etal-2021-nlp} and the MAWPS benchmark \citep{koncel-kedziorski-etal-2016-mawps}. We again account for the fact that we test all models in a zero-shot setup by using a more lenient evaluation criterion: As the required output is always a number, we simply check for the first number predicted by the model.\footnote {An exception to this is if the model's prediction contains an equation (e.g., ``The correct answer is 5+3=8''), in which case we consider the first number after the ``='' sign to be its prediction.}

Table~\ref{tab:math_results} shows results for all benchmarks. While GPT-J and GPT-J + CC perform about the same, \ours{} achieves stronger results even when API calls are disabled. We surmise that this is because the model is finetuned on many examples of API calls and their results, improving its own mathematical capabilities. Nonetheless, allowing the model to make API calls more than doubles performance for all tasks, and also clearly outperforms the much larger OPT and GPT-3 models. This is because across all benchmarks, for 97.9\% of all examples the model decides to ask the calculator tool for help.

\begin{table}
    \centering
    \small
    \setlength{\tabcolsep}{5pt}
    \begin{tabularx}{\linewidth}{Xccc}
        \toprule
        \textbf{Model} & \textbf{ASDiv} & \textbf{SVAMP} & \textbf{MAWPS} \\
        \midrule
         GPT-J & \phantom{0}7.5 & \phantom{0}5.2 & \phantom{0}9.9 \\         
         GPT-J + CC & \phantom{0}9.6 & \phantom{0}5.0 & \phantom{0}9.3 \\        
         \ours{} (disabled) & 14.8 & \phantom{0}6.3 & 15.0 \\
        \ours{} & \underline{\textbf{40.4}} & \underline{\textbf{29.4}} & \underline{\textbf{44.0}} \\
         \midrule
         OPT (66B) & \phantom{0}6.0 & \phantom{0}4.9 & \phantom{0}7.9 \\
         GPT-3 (175B) & 14.0 & 10.0 & 19.8 \\
         \bottomrule
    \end{tabularx}
    \caption{Results for various benchmarks requiring mathematical reasoning. \ours{} makes use of the calculator tool for most examples, clearly outperforming even OPT (66B) and GPT-3 (175B).}
    \label{tab:math_results}
\end{table}

\subsubsection{Question Answering}

We look at Web Questions \citep{berant-etal-2013-semantic}, Natural Questions \citep{kwiatkowski-etal-2019-natural} and TriviaQA \citep{joshi-etal-2017-triviaqa}, the three question answering datasets considered by \citet{brown2020language}. For evaluation, we check whether the first 20 words predicted by a model contain the correct answer instead of requiring an exact match. For \ours{}, we disable the question answering tool as this would make solving the tasks trivial, especially given that the underlying QA system was finetuned on Natural Questions.

Results are shown in Table~\ref{tab:qa_results}. Once again, \ours{} clearly outperforms all other models based on GPT-J, this time mostly relying on the Wikipedia search API (99.3\%) to find relevant information.  However, \ours{} still lags behind the much larger GPT-3 (175B) model. This is likely due to both the simplicity of our search engine (in many cases, it returns results that are clearly not a good match for a given query) and the inability of \ours{} to \emph{interact} with it, e.g., by reformulating its query if results are not helpful or by browsing through multiple of the top results. We believe that adding this functionality is an exciting direction for future work.

\begin{table}
    \centering
    \small
    \setlength{\tabcolsep}{5pt}
    \begin{tabularx}{\linewidth}{Xccc}
        \toprule
        \textbf{Model} & \textbf{WebQS} & \textbf{NQ} & \textbf{TriviaQA}  \\
        \midrule
         GPT-J & 18.5 & 12.8 & 43.9  \\         
         GPT-J + CC  & 18.4 & 12.2 & 45.6  \\        
        \ours{} (disabled)  & 18.9 & 12.6 & 46.7  \\
        \ours{} & \textbf{26.3} & \textbf{17.7} & \textbf{48.8} \\
         \midrule  
         OPT (66B) & 18.6 & 11.4 & 45.7 \\
         GPT-3 (175B) & \underline{29.0} & \underline{22.6} & \underline{65.9} \\
         \bottomrule
    \end{tabularx}
    \caption{Results for various question answering dataset. Using the Wikipedia search tool for most examples, \ours{} clearly outperforms baselines of the same size, but falls short of GPT-3 (175B).}
    \label{tab:qa_results}
\end{table}

\subsubsection{Multilingual Question Answering}
We evaluate \ours{} and all baseline models on MLQA~\citep{lewis2019mlqa}, a multilingual question-answering benchmark. A context paragraph for each question is provided in English, while the question can be in Arabic, German, Spanish, Hindi, Vietnamese, or Simplified Chinese. 
In order to solve the task, the model needs to be able to understand both the paragraph and the question, so it may benefit from translating the question into English. Our evaluation metric is the percentage of times the model's generation, capped at 10 words, contains the correct answer. 

Results are shown in Table~\ref{tab:mt_results_percentage}. Using API calls consistently improves \ours{}'s performance for all languages, suggesting that it has learned to make use of the machine translation tool. Depending on the language, this tool is used for 63.8\% to 94.9\% of all examples; the only exception to this is Hindi, for which the machine translation tool is used in only 7.3\% of cases. However, \ours{} does not consistently outperform vanilla GPT-J. This is mainly because for some languages, finetuning on CCNet deteriorates performance; this might be due to a distribution shift compared to GPT-J's original pretraining data.

OPT and GPT-3 perform surprisingly weak across all languages, mostly because they fail to provide an answer in English despite being instructed to do so. A potential reason for GPT-J not suffering from this problem is that it was trained on more multilingual data than both OPT and GPT-3, including the EuroParl corpus~\citep{koehn2005europarl, gao2020pile}. As an upper bound, we also evaluate GPT-J and GPT-3 on a variant of MLQA where both the context and the question are provided in English. In this setup, GPT-3 performs better than all other models, supporting our hypothesis that its subpar performance on MLQA is due to the multilingual aspect of the task. 

\begin{table}
    \centering
    \small
    \setlength{\tabcolsep}{3pt}
    \begin{tabularx}{\linewidth}{Xcccccc}
        \toprule
        \textbf{Model} & \textbf{Es} & \textbf{De} & \textbf{Hi}  & \textbf{Vi} & \textbf{Zh}  & \textbf{Ar} \\
        \midrule
         GPT-J & 15.2 & \textbf{\underline{16.5}} & \phantom{0}1.3 & \phantom{0}8.2 & \textbf{\underline{18.2}} & \phantom{0}\textbf{\underline{8.2}} \\         
         GPT-J + CC & 15.7 & 14.9 & \phantom{0}0.5 & \phantom{0}8.3 & 13.7 & \phantom{0}4.6 \\             
         \ours{} (disabled) & 19.8 & 11.9 & \phantom{0}1.2 & 10.1 & 15.0 & \phantom{0}3.1 \\     
         \ours{} & \textbf{\underline{20.6}} & 13.5 & \phantom{0}\textbf{\underline{1.4}} & \textbf{\underline{10.6}} & 16.8 & \phantom{0}3.7  \\     
         \midrule  
         OPT (66B) & \phantom{0}0.3 & \phantom{0}0.1 & \phantom{0}1.1 & \phantom{0}0.2 & \phantom{0}0.7 & \phantom{0}0.1 \\     
         GPT-3 (175B)  & \phantom{0}3.4 & \phantom{0}1.1 & \phantom{0}0.1 & \phantom{0}1.7 & 17.7 & \phantom{0}0.1 \\     
         \midrule  
         GPT-J (All En) & 24.3 & 27.0 & 23.9 & 23.3 & 23.1 & 23.6 \\     
         GPT-3 (All En)  & 24.7 & 27.2 & 26.1 & 24.9 & 23.6 & 24.0 \\     
         \bottomrule
    \end{tabularx}
    \caption{Results on MLQA for Spanish (Es), German (De), Hindi (Hi), Vietnamese (Vi), Chinese (Zh) and Arabic (Ar). 
    While using the machine translation tool to translate questions is helpful across all languages, further pretraining on CCNet deteriorates performance; consequently, \ours{} does not consistently outperform GPT-J. The final two rows correspond to models that are given contexts and questions in English.}
    \label{tab:mt_results_percentage}
\end{table}

\subsubsection{Temporal Datasets}

To investigate the calendar API's utility, we  evaluate all models on \textsc{TempLAMA} \citep{dhingra-etal-2022-time} and a new dataset that we call \textsc{Dateset}. \textsc{TempLAMA} is a dataset built from Wikidata that contains cloze queries about facts that change with time (e.g., ``Cristiano Ronaldo plays for \_\_\_'') as well as the correct answer for the years between 2010 and 2020. \textsc{Dateset}, described in Appendix~\ref{sec:dateset}, is also generated through a series of templates, but populated using a combination of random dates/durations (e.g., ``What day of the week was it 30 days ago?''). Critically, knowing the current date is required to answer these questions. For both tasks, we use the same evaluation as for the original LAMA dataset.

Results shown in Table~\ref{tab:temporal_results} illustrate that \ours{} outperforms all baselines for both \textsc{TempLAMA} and \textsc{Dateset}. However, closer inspection shows that improvements on \textsc{TempLAMA} can not be attributed to the calendar tool, which is only used for 0.2\% of all examples, but mostly to the Wikipedia search and question answering tools, which \ours{} calls the most. This makes sense given that
named entities in \textsc{TempLama} are often so specific and rare that even knowing the exact date alone would be of little help. The best course of action for this dataset -- first querying the calendar API to get the current date, and then querying the question answering system with this date -- is not only prohibited by our restriction of using at most one API call per example, but also hard to learn for \ours{} given that all API calls in its training data are sampled independently.

For \textsc{Dateset}, on the other hand, the considerable improvement of \ours{} compared to other models can be fully accredited to the calendar tool, which it makes use of for 54.8\% of all examples.

\begin{table}
    \centering
    \small
    \begin{tabularx}{\linewidth}{Xccc}
        \toprule
        \textbf{Model} & \textbf{\textsc{TempLAMA}} & \textbf{\textsc{Dateset}}\\
        \midrule
         GPT-J & 13.7 & \phantom{0}3.9 \\         
         GPT-J + CC  & 12.9 & \phantom{0}2.9   \\        
         \ours{} (disabled)  & 12.7 & \phantom{0}5.9 \\
         \ours{} & \textbf{\underline{16.3}} & \textbf{\underline{27.3}} \\
         \midrule  
         OPT (66B) & 14.5 & \phantom{0}1.3 \\
         GPT-3 (175B) & 15.5 & \phantom{0}0.8 \\
         \bottomrule
    \end{tabularx}
    \caption{Results for the temporal datasets. \ours{} outperforms all baselines, but does not make use of the calendar tool for \textsc{TempLAMA}.}
    \label{tab:temporal_results}
\end{table}

\subsection{Language Modeling} 
\label{section:language-modeling}

In addition to verifying improved performance on various downstream tasks, we also want to ensure that language modeling performance of \ours{} does not degrade through our finetuning with API calls. To this end, we evaluate our models on two language modeling datasets: WikiText \citep{merity2017pointer} and a subset of 10,000 randomly selected documents from CCNet \citep{wenzek-etal-2020-ccnet} that were not used during training. Perplexities of various models are shown in Table~\ref{tab:perplexities}. As one would expect, finetuning on CCNet leads to slightly improved performance on a different CCNet subset, but it slightly deteriorates performance on WikiText, presumably because the original pretraining data for GPT-J is more similar to WikiText than our randomly selected subset of CCNet. Most importantly, however, training on $\mathcal{C}^*$ (our dataset annotated with API calls) does not lead to an increase in perplexity compared to training on $\mathcal{C}$ when API calls are disabled at inference time.\footnote{We do not evaluate the perplexity of \ours{} with API calls enabled as computing the probability $p_M(x_t \mid x_1, \ldots, x_{t-1})$ of token $x_t$ given $x_1, \ldots, x_{t-1}$ would require marginalizing over all potential API calls that the model could make at position $t$, which is intractable.}

\begin{table}
    \centering
    \small
    \begin{tabularx}{\linewidth}{Xcc}
        \toprule
        \textbf{Model} & \textbf{WikiText} & \textbf{CCNet} \\
        \midrule
         GPT-J & \textbf{\phantom{0}9.9} & 10.6 \\         
         GPT-J + CC & 10.3 & \textbf{10.5} \\        
         \ours{} (disabled) & 10.3 & \textbf{10.5} \\
         \bottomrule
    \end{tabularx}
    \caption{Perplexities of different models on WikiText and our validation subset of CCNet. Adding API calls comes without a cost in terms of perplexity for language modeling without any API calls.}
    \label{tab:perplexities}
\end{table}

\subsection{Scaling Laws}
\label{section:scaling-laws}

\begin{figure*}
    \centering
    \includegraphics[width=\linewidth]{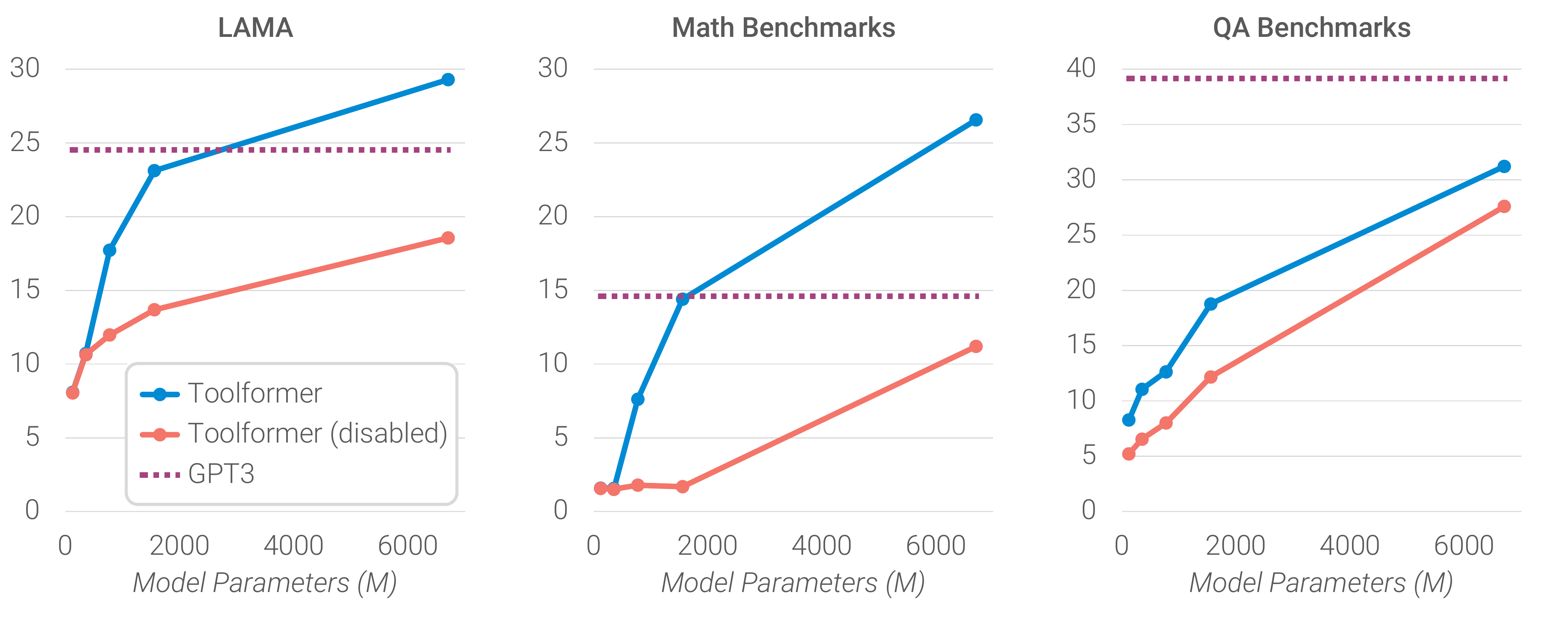}
    \caption{Average performance on LAMA, our math benchmarks and our QA benchmarks for GPT-2 models of different sizes and GPT-J finetuned with our approach, both with and without API calls. While API calls are not helpful to the smallest models, larger models learn how to make good use of them. Even for bigger models, the gap between model predictions with and without API calls remains high.}
    \label{fig:scaling_laws}
\end{figure*}

We investigate how the ability to ask external tools for help affects performance as we vary the size of our LM. To this end, we apply our approach not just to GPT-J, but also to four smaller models from the GPT-2 family \citep{radford2019language}, with 124M, 355M, 775M and 1.6B parameters, respectively. We do so using only a subset of three tools: the question answering system, the calculator, and the Wikipedia search engine. Apart from this, we follow the experimental setup described in Section~\ref{section:experimental-setup}.

Figure~\ref{fig:scaling_laws} shows that the ability to leverage the provided tools only emerges at around 775M parameters: smaller models achieve similar performance both with and without tools. An exception to this is the Wikipedia search engine used mostly for QA benchmarks; we hypothesize that this is because the API is comparably easy to use.
While models become better at solving tasks \emph{without} API calls as they grow in size, their ability to make good use of the provided API improves at the same time. As a consequence, there remains a large gap between predictions with and without API calls even for our biggest model.

\section{Analysis}
\label{section:analysis}

\paragraph{Decoding Strategy}

We investigate the effect of our modified decoding strategy introduced in Section~\ref{section:downstream-tasks}, where instead of always generating the most likely token, we generate the \texttt{<API>} token if it is one of the $k$ most likely tokens. Table~\ref{tab:top-k} shows performance on the T-REx subset of LAMA and on WebQS for different values of $k$. As expected, increasing $k$ leads to the model doing API calls for more examples -- from 40.3\% and 8.5\% with $k = 1$ (i.e., regular greedy decoding) to 98.1\% and 100\% for $k = 10$. While for T-REx, there is already a clear improvement in performance with greedy decoding, on WebQS our model only starts to make a substantial number of API calls as we slightly increase $k$. Interestingly, for $k = 1$ the model is calibrated to some extent: It decides to call APIs for examples that it would perform particularly badly on without making API calls. This can be seen from the fact that performance on examples where it decides \emph{not} to make an API call (44.3 and 19.9) is higher than average performance if no API calls are made at all (34.9 and 18.9). However, this calibration is lost for higher values of $k$.

\begin{table}
    \centering
    \small
    \newcolumntype{Y}{>{\centering\arraybackslash}X}
    \begin{tabularx}{\linewidth}{lYYYYlYYYc}
        \toprule
        \setlength{\tabcolsep}{1.2pt}
        & \multicolumn{4}{c}{\textbf{T-REx}} && \multicolumn{4}{c}{\textbf{WebQS}} \\
        \cmidrule{2-5}\cmidrule{7-10}
        $k$ & \textbf{All} & \textbf{AC} & \textbf{NC} & \textbf{\%} && \textbf{All} & \textbf{AC} & \textbf{NC} & \textbf{\%} \\
        \midrule
         0 & 34.9 & \phantom{0}-- & 34.9 & \phantom{0}0.0 && 18.9 & \phantom{0}-- & 18.9 & \phantom{10}0.0 \\         
         1 & 47.8 & 53.0 & 44.3 & 40.3 && 19.3 & 17.1 & 19.9 & \phantom{10}8.5 \\        
         3 & 52.9 & 58.0 & 29.0 & 82.8 && \textbf{26.3} & 26.5 & \phantom{0}6.6 & \phantom{1}99.3 \\
         10 & \textbf{53.5} & 54.0 & 22.5 & 98.1 && \textbf{26.3} & 26.4 & \phantom{0}-- & 100.0 \\
         \bottomrule
    \end{tabularx}
    \caption{\ours{} results on the T-REx subset of LAMA and on WebQS for different values of $k$ used during decoding. Numbers shown are overall performance (All), performance on the subset where the model decides to make an API call (AC) and all remaining examples (NC), as well as the percentage of examples for which the model decides to call an API (\%).}
    \label{tab:top-k}
\end{table}

\begin{table*}[ht]
    \renewcommand\tabularxcolumn[1]{m{#1}}
    \centering
    \small
    \begin{tabularx}{\linewidth}{Xcc}
    \toprule
    \textbf{Example} & $L_i^-\,{-}\,L_i^+$ & \textbf{Useful}  \\
    \midrule
    The Flodden Window (a war memorial dedicated to The Middleton Archers), in the Grade I-listed Church of St Leonard in Middleton is said to be the oldest war memorial in the United Kingdom. \texttt{<API>} \textbf{WikiSearch(War memorial Flodden) $\rightarrow$ Battle of Flodden > Commemoration > The stained-glass Flodden Window in Middleton Parish Church [\ldots] was constructed by Sir Richard Assheton in memory of the Battle of Flodden and the archers from Middleton who fought in it.} \texttt{</API>} Sir Richard Assheton of Middleton (who built St Leonard) was granted knighthood [\ldots] & 5.49 & \cmark \\
    \midrule
    Note: The WL will be open on Friday, \texttt{<API>} \textbf{Calendar() $\rightarrow$ Today is Thursday, March 9, 2017.} \texttt{</API>} March 10, and Sunday, March 19 for regular hours. & 2.11 & \cmark \\   
    \midrule
    The Nile has an approximate length of \texttt{<API>} \textbf{QA(What is the approximate length of the Nile?) $\rightarrow$ 6,853 km} \texttt{</API>} 6,853 kilometers, the White Nile being its main source. & 2.08 & \cmark \\
    \midrule
    If Venus had an atmosphere similar to Earth's then you would expect Venus' mean temperature to be 499 K (1.74 x 287) rather than 735 K which is \texttt{<API>} \textbf{Calculator(735 / 499) $\rightarrow$ 1.47} \texttt{</API>} 1.47 (735 / 499) times hotter than it should be. & 1.59 & \cmark \\
    \midrule
    You are here: Home / Featured / Catch this fast train to success! \texttt{<API>} \textbf{WikiSearch(Fast train success) $\rightarrow$ Fast Train > It also peaked at \#23 on the Canadian CHUM singles chart, on June 26, 1971. The success of this single established Myles Goodwyn as the band's main songwriter, and made it possible for April Wine to record a second album.} \texttt{</API>} Don’t wait weeks and $\left[\ldots\right]$ & 0.92 & \xmark \\
    \midrule
    Os Melhores Escolas em Jersey 2020 \texttt{<API>} \textbf{MT(Os Melhores Escolas em Jersey) $\rightarrow$ The Best Schools in Jersey} \texttt{</API>} On this page you can search for Universities, Colleges and Business schools in Jersey &  0.70 & \cmark \\
    \midrule
    Enjoy these pictures from the  \texttt{<API>} \textbf{Calendar() $\rightarrow$ Today is Friday, April 19, 2013.} \texttt{</API>} Easter Egg Hunt. & 0.33 & \cmark \\    
    \midrule
    85 patients (23\%) were hospitalised alive and admitted to a hospital ward. Of them, \texttt{<API>} \textbf{Calculator(85 / 23) $\rightarrow$ 3.70} \texttt{</API>} 65\% had a cardiac aetiology $\left[\ldots\right]$ & $-$0.02\phantom{$-$} & \xmark \\
    \midrule
    But hey, after the \texttt{<API>} \textbf{Calendar() $\rightarrow$ Today is Saturday, June 25, 2011.} \texttt{</API>} Disneyland fiasco with the fire drill, I think it’s safe to say Chewey won’t let anyone die in a fire. & $-$0.41\phantom{$-$} & \xmark \\ 
    \midrule
    The last time I was with \texttt{<API>} \textbf{QA(Who was last time I was with?) $\rightarrow$ The Last Time} \texttt{</API>} him I asked what he likes about me and he said he would tell me one day. & $-$1.23\phantom{$-$} & \xmark \\
    \bottomrule
    \end{tabularx}
    \caption{Examples of API calls for different tools, sorted by the value of $L_i^-\,{-}\,L_i^+$ that is used as a filtering criterion. High values typically correspond to API calls that are intuitively useful for predicting future tokens.
    }
    \label{fig:model_outputs}
\end{table*}

\paragraph{Data Quality} We qualitatively analyze some API calls generated with our approach for different APIs. Table~\ref{fig:model_outputs} shows some examples of texts from CCNet augmented with API calls, as well as the corresponding score $L_i^- - L_i^+$ that is used as a filtering criterion, and whether the API calls made by the model are intuitively useful in the given context. As can be seen, high values of $L_i^- - L_i^+$ typically correspond to useful API calls, whereas low values correspond to API calls that do not provide any information that is useful for predicting future tokens. There are some exceptions, e.g., an API call for ``Fast train success'' in the fourth example that does not give any relevant information but still reduces perplexity. However, some amount of noise in the API calls that are not filtered can actually be useful as it forces the model finetuned on $\mathcal{C}^*$ to not always blindly follow the results of each call it makes.

\section{Related Work}

\paragraph{Language Model Pretraining} There are various approaches that augment language models with some form of additional textual information during pretraining, including various forms of metadata \citep{keskar2019ctrl}, HTML tags \citep{aghajanyan2021htlm}, Wikipedia markup \citep{schick2022peer}, or related texts obtained from an information retrieval system \citep{guu2020realm,borgeaud2021retro,izacard2022atlas}. For all of these approaches, additional information is \emph{always} provided, regardless of whether it is helpful or not. In contrast, \ours{} learns for itself to explicitly asks for the right information.

\paragraph{Tool Use} Several approaches aim to equip LMs with the ability to use external tools such as search engines \citep{komeili-etal-2022-internet, thoppilan2022lamda, lazaridou2022internet,shuster2022blenderbot,yao2022react}, web browsers \citep{nakano2021webgpt}, calculators \citep{cobbe2021training,thoppilan2022lamda}, translation systems \citep{thoppilan2022lamda} and Python interpreters \citep{gao2022pal}. The way these models learn to use tools can roughly be divided into two approaches: Either they rely on large amounts of human supervision \citep{komeili-etal-2022-internet,nakano2021webgpt,thoppilan2022lamda} or they work by prompting the language model in a few-shot setup tailored towards a specific task where it is known a priori which tools needs to be used \citep{gao2022pal,lazaridou2022internet, yao2022react}. In contrast, the self-supervised nature of \ours{} enables it to learn how and when to use tools without requiring a specific prompt that shows task-specific examples of how a tool could be used. Perhaps most closely related to our work is TALM \citep{parisi2022talm}, an approach that uses a similar self-supervised objective for teaching a model to use a calculator and a search engine, but explores this only in settings where a model is finetuned for downstream tasks.

\paragraph{Bootstrapping} The idea of using self-training and bootstrapping techniques to improve models has been investigated in various contexts, ranging from word sense disambiguation \citep{yarowsky-1995-unsupervised}, relation extraction \citep{brin1999extracting,agichtein2000snowball}, parsing \citep{mcclosky-etal-2006-effective,reichart-rappoport-2007-self}, sequence generation \citep{He2020Revisiting}, few-shot text classification \citep{schick-schutze-2021-exploiting} and retrieval \citep{izacard2021distilling} to reasoning \citep{zelikman2022star}. In a similar spirit to these approaches, \ours{} is trained on its own predictions after applying a perplexity-based filtering step.

\section{Limitations}

While our approach enables LMs to learn how to use a variety of tools in a self-supervised way, there are some clear limitations to what can be achieved with our method in its current form. One such limitation is the inability of \ours{} to use tools in a \emph{chain} (i.e., using the output of one tool as an input for another tool). This is due to the fact that API calls for each tool are generated independently; as a consequence, there are no examples of chained tool use in the finetuning dataset. Our current approach also does not allow the LM to use a tool in an \emph{interactive} way -- especially for tools such as search engines, that could potentially return hundreds of different results, enabling a LM to browse through these results or to refine its search query in a similar spirit to \citet{nakano2021webgpt} can be crucial for certain applications. Beyond this, we found models trained with \ours{} to often be sensitive to the exact wording of their input when deciding whether or not to call an API; this is perhaps unsurprising given that LMs are known to be very sensitive to the prompt they are provided with in both zero-and few-shot settings \citep{jiang-etal-2020-know,schick-schutze-2021-exploiting}. Depending on the tool, our method is also very sample-inefficient; for example, processing more than a million documents results in only a few thousand examples of useful calls to the calculator API. A potential solution to this problem might be to iteratively apply our approach, similar to how this is done in related bootstrapping approaches \citep{schick-schutze-2021-exploiting,izacard2021distilling,parisi2022talm}. Finally, when deciding whether or not to make an API call, \ours{} currently does not take into account the tool-dependent, computational cost incurred from making an API call.

\section{Conclusion}

We have introduced \ours{}, a language model that learns in a self-supervised way how to use different tools such as search engines, calculators, and translation systems via simple API calls. This is done by finetuning on a large number of sampled API calls that are filtered based on whether they reduce perplexity on future tokens. \ours{} considerably improves zero-shot performance of a 6.7B parameter GPT-J model, enabling it to even outperform a much larger GPT-3 model on a range of different downstream tasks.

\bibliography{anthology,custom}
\bibliographystyle{acl_natbib}

\clearpage
\appendix

\section{API Details}
\label{appendix:api-details}

When sampling and filtering API calls, by default we use values of $\tau_s = 0.05$ and $\tau_f = 1.0$ -- i.e., we only make API calls at positions where the probability of the \texttt{<API>} token is at least 5\%, and we keep API calls if they reduce the loss by at least 1.0. We only keep the top $k = 5$ such positions and sample up to $m = 5$ API calls for each position identified in a piece of text. Due to the heuristic filtering described below, we generate API calls for the calculator and machine translation system on only a small subset of $\mathcal{C}$; to compensate for this, we set $\tau_s = 0.0$, $k = 20$ and $m = 10$ for these tools. As the resulting sets of API calls are still comparably small, we additionally set $\tau_f = 0.5$.

\subsection{Implementation}

\paragraph{Question Answering} We use the Atlas model of \citet{izacard2022atlas} finetuned on Natural Questions \citep{kwiatkowski-etal-2019-natural} as our question answering system. For creating $\mathcal{C}^*$ we use Atlas-large, enabling us to efficiently process millions of API calls; during inference, we use the larger Atlas-xxl model.

\paragraph{Calculator} Our calculator is based on a simple Python script and only supports the operators ``$+$'', ``$-$'', ``$*$'', and ``$/$''. It does not return any result for syntactically invalid equations. For sampling API calls, we apply heuristic filters to our subset of CCNet and only process documents that either (i) contain at least three numbers within a window of 100 tokens, where one of these numbers is the result of applying a mathematical operation to the other two, (ii) contain one of the sequences ``='', ``equals'', ``equal to'', ``total of'', ``average of'' followed by a number, or (iii) contain at least three numbers; for texts that only match the last criterion, we only keep a random subset of 1\%.

\paragraph{Calendar} For creating our dataset $\mathcal{C}^*$, we operate under the assumption that the calendar date in such cases should be the date that the document was created. We approximate this by extracting the date from the URL, if it is present. We filter out texts for which a date cannot be extracted, leaving around 18\% of the documents.

\paragraph{Machine Translation} For both training and inference, we use the 600M parameter NLLB~\citep{costa2022no} as our machine translation (MT) model. The source language is automatically detected using the fastText classifier~\citep{joulin2016fasttext}, while the target language is always set to English. Since most of the CCNet dataset is in English, we filter out the parts that contain only English text before generating API calls. More specifically, we only keep those paragraphs which contain text chunks in a language other than English preceded and followed by English text. We use text chunks of size 10 tokens. To determine whether the middle text chunk is in a language different than English we again use the fastText classifier with a confidence greater than 0.8. We also filter out any text chunks that contain only numbers or special symbols. This filtering mechanism allows us to generate data more efficiently by focusing our API call generations in places where the MT tool is likely to be helpful. After generating the MT API calls, we additionally remove from our training set those where the input to the MT tool appears after the API call but not before it. While during data generation the model can look ahead to generate API calls, this is not possible at inference time, so we want to dissuade the model from calling the API in such cases.

\subsection{Prompts}
\label{appendix:tool-prompts}

Below, we list the prompts used to sample API calls for each tool considered.

\paragraph{Question Answering} We use the following prompt for the question answering tool:
{\small
\begin{spverbatim}
Your task is to add calls to a Question Answering API to a piece of text. The questions should help you get information required to complete the text. You can call the API by writing "[QA(question)]" where "question" is the question you want to ask. Here are some examples of API calls:
Input: Joe Biden was born in Scranton, Pennsylvania.
Output: Joe Biden was born in [QA("Where was Joe Biden born?")] Scranton, [QA("In which state is Scranton?")] Pennsylvania.

Input: Coca-Cola, or Coke, is a carbonated soft drink manufactured by the Coca-Cola Company.
Output: Coca-Cola, or [QA("What other name is Coca-Cola known by?")] Coke, is a carbonated soft drink manufactured by [QA("Who manufactures Coca-Cola?")] the Coca-Cola Company.

Input: x
Output:
\end{spverbatim}}

\paragraph{Calculator} We use the following prompt for the calculator:
{\small
\begin{spverbatim}
Your task is to add calls to a Calculator API to a piece of text. The calls should help you get information required to complete the text. You can call the API by writing "[Calculator(expression)]" where "expression" is the expression to be computed. Here are some examples of API calls:
Input: The number in the next term is 18 + 12 x 3 = 54.
Output: The number in the next term is 18 + 12 x 3 = [Calculator(18 + 12 * 3)] 54.

Input: The population is 658,893 people. This is 11.4% of the national average of 5,763,868 people.
Output: The population is 658,893 people. This is 11.4% of the national average of [Calculator(658,893 / 11.4%)] 5,763,868 people.

Input: A total of 252 qualifying matches were played, and 723 goals were scored (an average of 2.87 per match). This is three times less than the 2169 goals last year.
Output: A total of 252 qualifying matches were played, and 723 goals were scored (an average of [Calculator(723 / 252)] 2.87 per match). This is twenty goals more than the [Calculator(723 - 20)] 703 goals last year.

Input: I went to Paris in 1994 and stayed there until 2011, so in total, it was 17 years.
Output: I went to Paris in 1994 and stayed there until 2011, so in total, it was [Calculator(2011 - 1994)] 17 years.

Input: From this, we have 4 * 30 minutes = 120 minutes.
Output: From this, we have 4 * 30 minutes = [Calculator(4 * 30)] 120 minutes.

Input: x
Output:
\end{spverbatim}}

\paragraph{Wikipedia Search} We use the following prompt for the Wikipedia search tool:
{\small
\begin{spverbatim}
Your task is to complete a given piece of text. You can use a Wikipedia Search API to look up information. You can do so by writing "[WikiSearch(term)]" where "term" is the search term you want to look up. Here are some examples of API calls:
Input: The colors on the flag of Ghana have the following meanings: red is for the blood of martyrs, green for forests, and gold for mineral wealth.
Output: The colors on the flag of Ghana have the following meanings: red is for [WikiSearch("Ghana flag red meaning")] the blood of martyrs, green for forests, and gold for mineral wealth.

Input: But what are the risks during production of nanomaterials? Some nanomaterials may give rise to various kinds of lung damage.
Output: But what are the risks during production of nanomaterials? [WikiSearch("nanomaterial production risks")] Some nanomaterials may give rise to various kinds of lung damage.

Input: Metformin is the first-line drug for patients with type 2 diabetes and obesity.
Output: Metformin is the first-line drug for [WikiSearch("Metformin first-line drug")] patients with type 2 diabetes and obesity.

Input: x
Output:
\end{spverbatim}}

\paragraph{Machine Translation} We use the following prompt for the machine translation tool:

{\small
\begin{spverbatim}
Your task is to complete a given piece of text by using a Machine Translation API.
You can do so by writing "[MT(text)]" where text is the text to be translated into English. 
Here are some examples:

Input: He has published one book: O homem suprimido (“The Supressed Man”)
Output: He has published one book: O homem suprimido [MT(O homem suprimido)] (“The Supressed Man”) 

Input: In Morris de Jonge’s Jeschuah, der klassische jüdische Mann, there is a description of a Jewish writer
Output: In Morris de Jonge’s Jeschuah, der klassische jüdische Mann [MT(der klassische jüdische Mann)], there is a description of a Jewish writer

Input: 南京高淳县住房和城乡建设局 城市新区设计 a plane of reference Gaochun is one of seven districts of the provincial capital Nanjing 
Output: [MT(南京高淳县住房和城乡建设局 城市新区设计)] a plane of reference Gaochun is one of seven districts of the provincial capital Nanjing 

Input: x
Output:
\end{spverbatim}}

\paragraph{Calendar} We use the following prompt for the calendar tool:

{\small
\begin{spverbatim}
Your task is to add calls to a Calendar API to a piece of text. The API calls should help you get information required to complete the text. You can call the API by writing "[Calendar()]" Here are some examples of API calls:

Input: Today is the first Friday of the year.
Output: Today is the first [Calendar()] Friday of the year.

Input: The president of the United States is Joe Biden.
Output: The president of the United States is [Calendar()] Joe Biden.

Input: The current day of the week is Wednesday. 
Output: The current day of the week is [Calendar()] Wednesday.

Input: The number of days from now until Christmas is 30. 
Output: The number of days from now until Christmas is [Calendar()] 30.

Input: The store is never open on the weekend, so today it is closed. 
Output: The store is never open on the weekend, so today [Calendar()] it is closed.

Input: x
Output:
\end{spverbatim}}

\section{Toolformer Training}
\label{appendix:finetuning}

We use up to 25k examples per API. Max sequence length 1,024. Effective batch size of 128.  All models are trained using DeepSpeed’s ZeRO-3 (Rasley
et al., 2020). We used 8 NVIDIA A100 40GB GPUs with BF16. Training up to 2k steps, where we evaluate PPL on a small development set from CCNet containing 1,000 examples every 500 steps. We pick the checkpoint that performs best. 

\section{Zero-Shot Prompts}

\subsection{LAMA and \textsc{TempLAMA}}

For both LAMA and \textsc{TempLAMA}, given an input text $\mathbf{x}$, we use the following prompt: \texttt{Please complete the following text so that it is factually correct: $\mathbf{x}$}.

\subsection{Math Benchmarks}

For all math benchmarks, given a context $\mathbf{x}$ and a question $\mathbf{q}$, our prompt is: $\mathbf{x}\ \mathbf{q}$ \texttt{The answer is}.

\subsection{Question Answering}

For all question answering datasets, including \textsc{Dateset}, we simply prefix the question with \texttt{Answer the following question:}. We append a question mark if the question does not already end with one.

\subsection{Multilingual Question Answering}

For MLQA, given a context $\mathbf{x}$ and a question $\mathbf{q}$, our prompt is: \texttt{Your task is to answer a question based on the following paragraph: $\mathbf{x}$ Now answer the following question in English: $\mathbf{q}$}.

\section{\textsc{Dateset}}
\label{sec:dateset}
\textsc{Dateset} is created by first randomly selecting 500 ``current dates''. For each current date, another relatively past/future date is randomly selected within a four-year range, and the two dates are used to fill the query templates in Table~\ref{tab:dateset_stats}. An example of one such query using the first template would be, ``How many days ago was August 14, 2020?'' If called, the Calendar tool would return the presumed current date (e.g., ``Today is Sunday, November 20, 2020''). 

\def\arraystretch{1.5}
\begin{table}
    \centering
    \small
    \setlength{\tabcolsep}{5pt}
    \begin{tabularx}{\linewidth}{Xcc}
        \toprule
        \textbf{Template} & \textbf{Size}\\
        \midrule     
        How many days \{ago was, are there until\} \{\textit{past\_date}, \textit{future\_date\}}?  & \phantom{00,}400\\
        
        What \{day of the week, day of the month, month, year\} was it (\textit{current\_date -- past\_date}) \{days, weeks, months, years\} ago?  & \phantom{00,}800\\

         What \{day of the week, day of the month, month, year\} will it be in (\textit{future\_date -- current\_date}) days?  & \phantom{00,}800\\
        What day of the week \{is, was\} it on \{\textit{past\_date}, \textit{future\_date\}}? & \phantom{00,}400\\ 
        What \{day of the week, day of the month, month, year\} \{is, was\} it \{the day before yesterday, yesterday, today, tomorrow, the day after tomorrow\}? & \phantom{0}4,000\\
        What \{day of the week, day of the month, month\} \{is, was\} $holiday$ this year? & \phantom{0}1,800\\
        How many \{days, weeks, months, years\} \{ago was, are there until\} $holiday$ this year?  & \phantom{0}1,200\\
        \midrule
        Total & \phantom{0}9,400 \\
         \bottomrule
    \end{tabularx}
    \caption{Templates used to create \textsc{Dateset} where a \textit{current\_date} is randomly selected. For each \textit{current\_date}, a random \textit{past\_date} and \textit{future\_date} is generated and used to fill each template, if relevant. The federal holidays in the United States (e.g., Thanksgiving) were used in the templates involving holidays.}
    \label{tab:dateset_stats}
\end{table}

\end{CJK*}
\end{document}